\pdfoutput=1

\documentclass[11pt]{article}

\usepackage[final]{acl}
\usepackage{times}
\usepackage{latexsym}

\usepackage[T1]{fontenc}

\usepackage[utf8]{inputenc}
\usepackage{booktabs}
\usepackage{microtype}
\usepackage{arydshln}
\usepackage{inconsolata}
\usepackage{dashrule}

\usepackage{graphicx}
\usepackage{kotex}
\usepackage{arydshln}
\usepackage{booktabs}
\usepackage{multirow}
\usepackage{hyperref}
\usepackage{colortbl}
\usepackage{amssymb}
\usepackage{xcolor}
\usepackage{amsmath}
\usepackage{arydshln}
\usepackage{algorithm}
\usepackage {algpseudocode}
\usepackage{algorithmicx}
\usepackage{algcompatible}
\usepackage[colorinlistoftodos]{todonotes}

\newcommand\blfootnote[1]{%
  \begingroup
  \renewcommand\thefootnote{}\footnote{#1}%
  \addtocounter{footnote}{-1}%
  \endgroup
}

\title{\textsc{Revise}: A Framework for Revising OCRed text in Practical Information Systems with Data Contamination Strategy}

\author{
 \textbf{Gyuho Shim\textsuperscript{1$\ast$}},
 \textbf{Seongtae Hong\textsuperscript{1$\ast$}},
 \textbf{Heuiseok Lim\textsuperscript{1,2}$^{\dagger}$}
\\
 \textsuperscript{1}Department of Computer Science and Engineering, Korea University\\
 \textsuperscript{2}Human-inspired AI Research,
\\
\text{\{gjshim, ghdchlwls123, limhseok\}@korea.ac.kr}}

\begin{document}
\maketitle
\begin{abstract}
Recent advances in Large Language Models (LLMs) have significantly improved the field of Document AI, demonstrating remarkable performance on document understanding tasks such as question answering. However, existing approaches primarily focus on solving specific tasks, lacking the capability to structurally organize and manage document information. To address this limitation, we propose \textsc{Revise}, a framework that systematically corrects errors introduced by OCR at the character, word, and structural levels. Specifically, \textsc{Revise} employs a comprehensive hierarchical taxonomy of common OCR errors and a synthetic data generation strategy that realistically simulates such errors to train an effective correction model. Experimental results demonstrate that \textsc{Revise} effectively corrects OCR outputs, enabling more structured representation and systematic management of document contents. Consequently, our method significantly enhances downstream performance in document retrieval and question answering tasks, highlighting the potential to overcome the structural management limitations of existing Document AI frameworks.

\end{abstract}
\blfootnote{$\ast$ Equal contributions}
\blfootnote{$\dagger$ Co-corresponding author}

\section{Introduction}
\begin{figure}[hbt!]
\centering 
\includegraphics[width=0.78\linewidth]{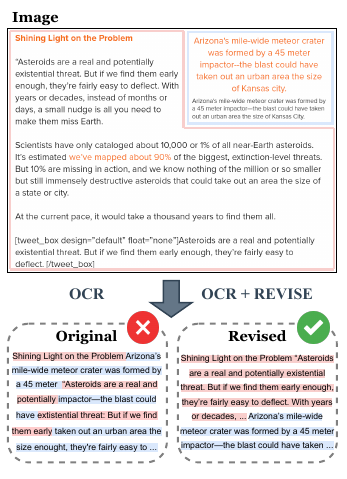}
\caption{Illustration comparing conventional OCR and OCR+\textsc{Revise} processing in a multi-column setting. \textit{Left}: text conflation with merged topics. \textit{Right}: \textsc{Revise} reconstructs separate textual elements into properly structured content.}
\label{fig:illustrate} 
\end{figure}

Recent advances in Natural Language Processing (NLP), particularly with Large Language Models (LLMs)~\cite{minaee2024largelanguagemodelssurvey}, have demonstrated remarkable performance on core tasks such as Question Answering (QA), reasoning and Retrieval Augmented Generation (RAG)~\cite{gao2024retrievalaugmentedgenerationlargelanguage}, thereby substantially broadening their formidable applicability. Moreover, recent research has rapidly expanded towards Document AI, aiming to understand and effectively utilize structured and complex information within real-world documents~\cite{cui2021documentaibenchmarksmodels,hong-etal-2024-intelligent}.

In particular, there is increasing interest in leveraging text extracted via Optical Character Recognition (OCR)~\cite{subramani2021surveydeeplearningapproaches} and document analysis techniques, along with layout information obtained from original documents, to enable LLMs to perform tasks over documents. However, current approaches have primarily focused on specific document understanding tasks~\cite{barboule2025surveyquestionansweringvisually}, leaving the broader goal of effectively preserving original document structure and converting documents into structured assets or databases underexplored. Typically, extracting and storing textual information from image-based documents requires OCR, which inevitably introduces recognition errors due to various factors, such as diverse fonts, deteriorated print quality, and layout complexities. Consequently, employing a simplistic processing pipeline for indexing or retrieving erroneous OCR text often leads to degraded performance. To effectively facilitate these applications, denoising OCR errors remains a critical prerequisite, necessitating a more sophisticated and resilient pipeline in Document AI.

In this paper, we propose \textsc{Revise}, designed to effectively address common OCR errors and accurately restore textual content while preserving the original document structure. To overcome the scarcity of high-quality annotated datasets for OCR error correction, we generate synthetic data using a realistic error injection methodology, in which diverse error patterns are systematically introduced into publicly available datasets. By training over these synthetic datasets, our model can effectively learn representative OCR errors and robustly reconstruct documents in their original forms, thereby enabling the accurate preservation and storage of textual information. Experimental evaluations on downstream tasks, including retrieval and question answering, further demonstrate that \textsc{Revise} maintains strong performance even without explicit OCR-error-correction annotations, showing broad applicability across various document types. Our contributions are as follows:

\begin{itemize} 
\item Systematically analyzes and categorizes error types frequently encountered in OCR-based real-world document processing scenarios.

\item Proposes \textsc{Revise}, an effective revision method leveraging synthetic datasets created by realistically emulating error patterns in publicly available datasets.

\item Demonstrates through extensive experiments that \textsc{Revise} significantly improves document retrieval and question answering, while substantially enhancing semantic coherence and readability.

\end{itemize}

\section{Related Works}

\subsection{Optical Character Recognition}
OCR serves as a foundation of document digitization, transforming images and scanned documents into searchable digital content \cite{article_OCRfoundation}. At its core, CNNs and RNNs are employed to recognize visual patterns in document images and convert them to text \cite{RNN_OCR_CHEN, vinyals2015show, qiang2016memorymattersconvolutionalrecurrent, 6126402, 6460871}, with tools like Tessearct \cite{4376991} and EasyOCR\footnote{\url{https://github.com/JaidedAI/EasyOCR}} in widespread use. Modern systems often utilize encoder-decoder architectures with attention mechanisms to improve recognition accuracy \cite{kim2022ocrfreedocumentunderstandingtransformer}. 

Despite advancements, OCR systems face limitations with image quality and complex layouts. Errors induced from such issues propagate to downstream applications: in information retrieval, studies \cite{4624283, 10.1007/s00799-023-00345-6, 10.1007/978-3-030-45442-5_13, zhang2025ocrhindersragevaluating} have demonstrated that OCR errors substantially degrade retrieval performance by transforming valid words into misspellings that impact term frequencies and relevance scoring. Additionally, OCR errors significantly impact document reasoning tasks \cite{gupte2021lightscameraactionframework, artidigh20, in_depth_analysis_of_OCR}, with extensive research showing cascading effects on document understanding and knowledge base construction, as entities and relationships extracted from OCR text often contain errors that compound through subsequent processing steps, ultimately compromising the reliability of AI systems that are contingent upon accurate document content.

\subsection{Document AI Methods}
Document AI applies AI techniques to understand, process, and extract information from document images \cite{cui2021documentaibenchmarksmodels}, focusing on four main tasks: Document Layout Analysis \cite{zhong2019publaynetlargestdatasetdocument, li2020docbankbenchmarkdatasetdocument}, Document Visual Question Answering \cite{mathew2021docvqa, VisualMRC2021, chen2021websrcdatasetwebbasedstructural}, Visual Information Extraction \cite{8977955, wang2021robustvisualinformationextraction, park2019cord}, and Document Image Classification \cite{harley2015evaluationdeepconvolutionalnets, 19694}. To address OCR shortcomings while excelling at these tasks, two major paradigms have emerged in Document AI.

The first approach involves OCR-free Multimodal LLMs \cite{huang2022layoutlmv3pretrainingdocumentai, liu2024textmonkeyocrfreelargemultimodal, li2021structextstructuredtextunderstanding, kim2022ocrfreedocumentunderstandingtransformer}, which process images directly without explicit text extraction. These models achieve impressive performance in document understanding and reasoning through vision-language pretraining; however, their reliance on extensive annotated datasets and computationally intensive training poses considerable challenges for practical deployment, especially in resource-constrained scenarios. The second approach integrates OCR-based LLMs \cite{perot2024lmdxlanguagemodelbaseddocument, he2023icld3ieincontextlearningdiverse, wang2023docllmlayoutawaregenerativelanguage, lu2024boundingboxworthtoken}, extracting text via OCR before applying an LLM for reasoning. While leveraging existing OCR technology, this approach inherits OCR errors and focuses primarily on reasoning-based tasks like question answering and information extraction.

Existing approaches exhibit task dependency, prioritizing answering and reasoning but neglecting crucial intermediate steps like assetization for information retrieval. Our method addresses this issue by providing a task-independent framework, enabling structured OCR outputs that can be effectively utilized in databases or knowledge bases.

\section{\textsc{Revise}}

The \textsc{Revise} framework systematically addresses OCR errors that occur at the character, word, and structural levels. Specifically, our approach involves: (1) a comprehensive OCR error taxonomy that hierarchically categorizes errors according to their linguistic granularities, (2) a contamination strategy for synthesizing realistic error patterns by injecting them into clean datasets, and (3) a training procedure designed to revise contaminated text sequences back to their original forms.

\subsection{OCR Error Categorization}
\label{categorize}

OCR errors negatively impact various downstream NLP tasks, including key extraction, named entity recognition, and information retrieval. \citet{error_cascade} has demonstrated that errors introduced in early processing stages propagate to subsequent stages, resulting in cumulative error cascades.
Motivated by these challenges, we conduct a comprehensive analysis of OCR error patterns across various document types. Based on the scope and influence of errors within textual structures, we propose a hierarchical OCR error taxonomy as illustrated with examples in Table~\ref{tab:categorization}, consistent with existing frameworks found in the post-OCR correction literature.

\begin{table}[!ht]
\centering
\resizebox{1\columnwidth}{!}{
\begin{tabular}{l l p{4.2cm}}
\toprule
\textbf{Category} & \textbf{Name} & \textbf{Example} \\
\midrule

\multirow{7}{*}{\shortstack[l]{\textbf{Character Level}\\(Single-character)}} 
  & Insertion 
    & apple → applee \\
  \cmidrule(lr){2-3}
  & \multirow{2}{*}{Deletion}
    & clamp → lamp \\
  & 
    & filter → filer \\
  \cmidrule(lr){2-3}
  & \multirow{2}{*}{Substitution}
    & O → 0, é → e \\
  & 
    & blue → b1ue \\
  \cmidrule(lr){2-3}
  & Transposition 
    & Gauge → Guage \\
\midrule

\multirow{2}{*}{\shortstack[l]{\textbf{Word Level}\\(Word-segmentation)}} 
  & Over-Segmentation
    & greenhouse → green house \\
  \cmidrule(lr){2-3}
  & Under-Segmentation
    & Not able → Notable \\
\midrule

\multirow{2}{*}{\shortstack[l]{\textbf{Column Level}\\(Layout-reading)}}
  & \multirow{2}{*}{Column Reading Order}
  & \multirow{2}{*}{\textbf{Figure~\ref{fig:illustrate}}} \\
& & \\ 
\bottomrule
\end{tabular}
}
\caption{OCR Error Categorization}
\label{tab:categorization}
\end{table}

\subsubsection*{Character-level}
Character-level errors encompass a range of misrecognitions and distortions that occur at the individual character scale, fundamentally altering the basic building blocks of text and potentially cascading into more significant semantic disruptions. \textbf{\textit{Insertion}} represents the addition of spurious characters into the text stream, commonly resulting from document noise, artifacts, or scanner interference~\cite{afli-etal-2016-using, kashid2025roundtripocrdatagenerationtechnique}. \textbf{\textit{Deletion}} involves the omission of legitimate characters, frequently occurring when poor contrast or faded text prevent proper recognition~\cite{7991582}. \textbf{\textit{Substitution}} occurs when the OCR incorrectly identifies characters, replacing them with visually similar alternatives due to font peculiarities or resolution limitations, resulting in common confusions such as ``l/1/!'',``5/S'' and ``0/O''~\cite{artidigh20, essay102117}. \textbf{\textit{Transposition}} results in character position swapping, often stemming from bounding box coordinate miscalculations~\cite{suissa2023optimizingneuralnetworktraining}.

\subsubsection*{Word-level} 
Word-level errors primarily manifest as improper segmentation issues, where the boundaries between words are incorrectly identified, leading to the fragmentation or merging of terms and significantly impacting the lexical integrity of the processed text. Segmentation stems from OCR's misidentification of word boundaries, taking the form of two distinct types~\cite{suissa2023optimizingneuralnetworktraining, afli-etal-2016-using}. \textbf{\textit{Over-segmentation}} occurs when OCR incorrectly inserts word boundaries (i.e., extra space) within what should be a single word, fragmenting cohesive terms into separate components. \textbf{\textit{Under-segmentation}} results from distinct words erroneously combining into a single unit due to spacing misinterpretation or layout analysis failures. \citet{nastase-hitschler-2018-correction} demonstrate how these errors impact keyword extraction and information retrieval, as they alter token distribution and disrupt phrase-level semantics.

\subsubsection*{Column-level}
Column-level errors refer to structural misinterpretations that disrupt the logical flow of text and distort the intended document layout. Documents with multiple columns are particularly vulnerable to these errors, potentially misarranging reading order and weakening overall coherence and readability. \textbf{\textit{Column reading order}} frequently arises due to the common assumption of a standard reading order from left to right and top to bottom. This assumption tends to cause incorrect interpretations of logical continuity within multi-column layouts, leading to misplaced text segments~\cite{wang2023textreadingorderuncontrolled, wang2021layoutreaderpretrainingtextlayout}. Such layout errors can significantly impact various downstream NLP tasks, severely compromising overall task performance even when the OCR's textual output itself is relatively accurate~\cite{artidigh20}.

By categorizing OCR errors according to this hierarchical taxonomy, it becomes possible to devise customized correction strategies tailored to tackle specific errors at their corresponding levels of textual organization. This approach serves as a foundation for generating effective error revision datasets.

\subsection{Data Contamination Strategy}
\label{strategy}
To train the revision model effectively, we utilize publicly available datasets and systematically introduce synthetic OCR errors based on the error categories defined in Section~\ref{categorize}. Our contamination strategy is designed to mimic both structural and granular OCR failures in a controlled manner, creating a realistic training corpus that reflects the hierarchical error patterns observed in real-world OCR outputs.

The contamination process unfolds in two stages. First, we create a structured template by dividing the raw text into fixed-length lines, reformatting to a single column layout. Next, we simulate \textit{Column reading order} errors by segmenting the text into sections, converting selected sections into multi-column formats, and reading horizontally across columns instead of vertically down each column. This approach mirrors how OCR systems typically misinterpret multi-column layouts, where text is incorrectly read left to right across columns rather than processing each column separately.

In the second stage, after the structural reordering, a set of error functions is applied to introduce distortions at the character, word, and sentence levels. \textit{Deletion}, \textit{Insertion}, \textit{Substitution}, and \textit{Transposition} are applied probabilistically, while \textit{Segmentation} errors are introduced by either inserting extra spaces or omitting existing spaces. Each error function is governed by configurable parameters to ensure a realistic blend of error types. The framework supports multiple contamination settings;
 
in this work, we primarily adopt a configuration that emphasizes fine-grained perturbations. This approach closely emulates common OCR errors while maintaining sufficient overall document coherence. {Detailed information regarding the contamination algorithms and parameter ratios can be found in the Appendix~\ref{apx:cp}.} 
The final output is a contaminated corpus reflecting typical OCR-induced distortions, forming the basis for training our \textsc{Revise} model to correct OCR outputs and improve downstream document processing tasks robustly.

\begin{table*}[t!]
\centering
\renewcommand{\arraystretch}{1}
\setlength{\tabcolsep}{9pt}

\resizebox{0.95\textwidth}{!}
{
\begin{tabular}{lccc ccc ccc ccc c}
\toprule
 \multirow{2.3}{*}{Methods}  & \multicolumn{3}{c}{bge-large-en-v1.5} & \multicolumn{3}{c}{e5-large-v2} & \multicolumn{3}{c}{jina-embeddings-v2-base} & \multicolumn{3}{c}{gte-base-en-v1.5} & \multirow{2}{*}{Avg} \\
\cmidrule(lr){2-4} \cmidrule(lr){5-7} \cmidrule(lr){8-10} \cmidrule(lr){11-13}
 & @1 & @3 & @5 & @1 & @3 & @5 & @1 & @3 & @5 & @1 & @3 & @5 &  \\

\midrule
\multicolumn{14}{c}{VisualMRC}\\
\midrule

Baseline        
    & 0.5690 & 0.6928 & 0.7314
    & 0.6044 & 0.7208 & 0.7533
    & 0.5243 & 0.6418 & 0.6843
    & 0.5604 & 0.6859 & 0.7248
    & 0.6578 (6)\\\hline
\textsc{Revise}$_{meta}$         
    & 0.5793 & \textbf{0.7030} & 0.7422
    & 0.6076 & \textbf{0.7232} & 0.7592
    & 0.5352 & \textbf{0.6553} & 0.6951
    & 0.5696 & \textbf{0.6960} & \textbf{0.7336}
    & \textbf{0.6666} (1) \\

\textit{only} Column
    & 0.5751 & 0.6981 & 0.7348
    & 0.6005 & 0.7174 & 0.7539
    & 0.5306 & 0.6477 & 0.6868
    & 0.5665 & 0.6914 & 0.7321
    & 0.6612 (3)\\
\textit{only} Deletion
    & 0.5684 & 0.6910 & 0.7317
    & 0.5997 & 0.7190 & 0.7546
    & 0.5195 & 0.6404 & 0.6789
    & 0.5555 & 0.6856 & 0.7218
    & 0.6555 (8)\\
\textit{only} Insertion
    & 0.5687 & 0.6920 & 0.7303
    & 0.5991 & 0.7187 & 0.7524
    & 0.5233 & 0.6386 & 0.6828
    & 0.5578 & 0.6831 & 0.7220
    & 0.6557 (7)\\
\textit{only} Substitution
    & 0.5716 & 0.6936 & 0.7332
    & 0.6018 & 0.7196 & 0.7555
    & 0.5265 & 0.6430 & 0.6847
    & 0.5629 & 0.6869 & 0.7250
    & 0.6587 (4)\\
\textit{only} Segmentation
    & \textbf{0.5796} & 0.7021 & \textbf{0.7427}
    & \textbf{0.6078} & 0.7223 & \textbf{0.7612}
    & \textbf{0.5362} & 0.6515 & \textbf{0.6954}
    & \textbf{0.5719} & 0.6948 & 0.7323
    & 0.6665 (2)\\
\textit{only} Transposition
    & 0.5732 & 0.6938 & 0.7320
    & 0.6024 & 0.7169 & 0.7537
    & 0.5261 & 0.6440 & 0.6856
    & 0.5605 & 0.6884 & 0.7242
    & 0.6584 (5)\\

\midrule
\multicolumn{14}{c}{DUDE}\\
\midrule

Baseline        
    & 0.2013
    & 0.3087
    & 0.3490
    & 0.2013
    & 0.2718
    & 0.3188
    & 0.1342 
    & 0.1846
    & 0.2584
    & 0.2047 
    & 0.2886
    & 0.3188
    & 0.2534 (8)\\\hline
    
\textsc{Revise}$_{meta}$           
    & \textbf{0.2282}
    & 0.3121
    & 0.3523
    & 0.2248
    & 0.2987
    & 0.3255
    & \textbf{0.1980}
    & \textbf{0.2819}
    & \textbf{0.3221}
    & \textbf{0.2315}
    & \textbf{0.3121}
    & \textbf{0.3591}
    & 0.2975 (3)\\

\textit{only} Column
    & 0.2215
    & \textbf{0.3322}
    & \textbf{0.3691}
    & 0.2148
    & \textbf{0.3221}
    & \textbf{0.3792}
    & 0.1812
    & 0.2785
    & 0.3154
    & 0.2282
    & 0.3020 
    & 0.3423
    & \textbf{0.3076} (1)\\
    
\textit{only} Deletion
    & 0.1946
    & 0.2953
    & 0.3289
    & 0.2215
    & 0.2919
    & 0.3289
    & 0.1779
    & 0.255
    & 0.2886
    & 0.2047
    & 0.2987
    & 0.3423
    & 0.2729 (7)\\
    
\textit{only} Insertion
    & 0.1913
    & 0.2953
    & 0.3456
    & 0.198
    & 0.2819
    & 0.3054
    & 0.1309
    & 0.1711 
    & 0.2617
    & 0.1846
    & 0.2987
    & 0.3423
    & 0.2774 (5)\\
    
\textit{only} Substitution
    & 0.2013
    & 0.2987
    & 0.3456
    & 0.2047
    & 0.2819
    & 0.3221
    & 0.1913
    & 0.2852
    & 0.3087
    & 0.2215
    & 0.3020
    & 0.3356
    & 0.2819 (4)\\
    
\textit{only} Segmentation
    & 0.2215
    & 0.3087
    & 0.3658
    & \textbf{0.2483}
    & 0.3020 
    & 0.3389
    & 0.1779
    & 0.2349
    & 0.2886
    & 0.2517
    & 0.3054
    & 0.3322
    & 0.2987 (2)\\
    
\textit{only} Transposition
    & 0.1846
    & 0.2987
    & 0.3423
    & 0.198
    & 0.2718
    & 0.3188
    & 0.1779
    & 0.2383
    & 0.2886
    & 0.2181 
    & 0.2886
    & 0.3356
    & 0.2752 (6)\\
\bottomrule
\end{tabular}
}

\caption{Retrieval performance on VisualMRC and DUDE datasets using Recall@k (ranks in parentheses; best scores are in \textbf{bold})}
\label{tab:retrieval}
\end{table*}
\subsection{Training}

For effective OCR error correction, we design a total of seven \textsc{Revise} models, consisting of one main model trained comprehensively on all error types and six auxiliary models, each specialized individually on a specific error type. All models share an identical backbone architecture, the Llama-3.1-1B-Instruct~\footnote{\url{https://huggingface.co/meta-llama/Llama-3.2-1B-Instruct}}, and are trained on synthetic data generated using text sampled from the Wikipedia~\footnote{\url{https://huggingface.co/datasets/wikimedia/wikipedia}} corpus. To ensure fair and consistent comparisons between models, each dataset comprises an equal number of samples, totaling 30,000 data points.

The central model proposed in this paper, \textsc{Revise}$_{meta}$, is designed to robustly handle realistic and general document processing scenarios. Specifically, based on the strategy described in \S~\ref{strategy}, \textsc{Revise}$_{meta}$ is trained comprehensively on data that incorporates the six major error categories frequently confronted in practical OCR systems: column reading order, segmentation, deletion, substitution, insertion, and transposition errors. Thus, the model is capable of effectively handling and correcting complex and diverse errors that commonly arise during OCR processing of documents.

To precisely analyze the performance of \textsc{Revise} and to better understand the characteristics and correction difficulties associated with each error type, we further train six specialized auxiliary models, each focusing exclusively on a single type of OCR error. These specialized models are individually trained on data injected with only one specific error category, thereby allowing each model to be optimized for correcting its particular error type.

Through this experimental design, we evaluate the overall effectiveness and practical applicability of the \textsc{Revise}$_{meta}$ model when dealing with realistic OCR error scenarios. Additionally, comparisons between the generalized and respective error-targeted models enable us to quantify and analyze the relative importance and characteristics of each specific type of error, as well as their influence on the overall OCR error correction pipeline. Ultimately, our goal is to clearly identify the strengths and weaknesses of generalized versus error-specific approaches, dependent upon the characteristics of documents and distributions of errors encountered, thereby providing practically useful guidelines for real-world implementations.

\section{Experimental Setup}
\subsection{Models}

We evaluate the effectiveness of our proposed \textsc{Revise} framework on downstream tasks by employing embedding models and LLMs. For document retrieval, we adopt four recent embedding models: bge-large-en-v1.5~\cite{bge_embedding}, intfloat/e5-large-v2~\cite{wang2022text}, jina-embeddings-v2-base-en~\cite{gunther2023jina}, and gte-base-en-v1.5~\cite{li2023gte}. These models enable us to quantify how effectively OCR-corrected documents can be matched to queries. For question answering, we utilize two large instruction-tuned language models: Gemma-2-2b-it~\cite{gemma_2024} and Llama-3.1-8B-Instruct~\cite{meta_llama_2024}. By leveraging these models, we assess the capability of our correction method to enhance structured document comprehension and reasoning performance.

\subsection{Evaluation}

The performance of the proposed framework is evaluated on document Visual Question Answering (VQA) and Visual Information Extraction (VIE) datasets, focusing on three main aspects and comparing results between original OCR-extracted text and the text post-processed by \textsc{Revise}. First, we directly assess document retrieval performance using Recall@K (k=1,3,5) on the VisualMRC~\cite{VisualMRC2021} and DUDE~\cite{vanlandeghem2023documentunderstandingdatasetevaluation} datasets. Second, for DocVQA~\cite{mathew2021docvqa}, CORD~\cite{park2019cord}, and FUNSD~\cite{jaume2019funsddatasetformunderstanding}, we evaluate the textual similarity between documents and questions via BERTScore~\cite{zhang2020bertscoreevaluatingtextgeneration}\footnote{For DocVQA, CORD, and FUNSD datasets, pure IR-based metrics alone are insufficient to accurately measure performance due to duplicate questions and similar keywords; hence, we use textual similarity measures.}. Lastly, we compare QA performance of models on original OCR text versus REVISE-enhanced texts using standard evaluation metrics commonly used for each dataset: CIDEr~\cite{vedantam2014cider} for generative answer quality on VisualMRC and F1-score for answering performance on CORD.

\section{Experimental Results}

\begin{table}[t!]
\renewcommand{\arraystretch}{0.98}
\centering
\resizebox{0.83\columnwidth}{!}{
\begin{tabular}{lccc}
\toprule
Category & DocVQA & CORD & FUNSD \\
\midrule
Baseline  & 0.4959 (7) & 0.5390 (5) & 0.5577 (6) \\
\hline
\textsc{Revise}$_{meta}$ & \textbf{0.5137} (1)& \textbf{0.5443} (1)    & \textbf{0.5647} (1)\\
\textit{only} Column   & 0.4849 (8)& 0.5361 (6)& 0.5620 (2)\\
\textit{only} Deletion & 0.4960 (6)& 0.5346 (7)& 0.5603 (3)\\
\textit{only} Insertion & 0.5019 (3)& 0.5390 (5)& 0.5538 (8)\\
\textit{only} Substitution  & 0.4992 (5)& 0.5402 (3)& 0.5566 (7)\\
\textit{only} Segmentation  & 0.5096 (2)& 0.5408 (2)& 0.5601 (4)\\
\textit{only} Transposition & 0.5008 (4)& 0.5398 (4)& 0.5583 (5)\\
\bottomrule
\end{tabular}
}
\caption{BERTScore performance on query–document pairs for DocVQA, CORD, and FUNSD}
\label{bertscore}
\end{table}


\subsection{Understanding Evaluation}

\paragraph{Retrieval Performance}

Table \ref{tab:retrieval} presents a comparative analysis of various OCR error revisions and their impact on embedding-based text retrieval performance using the VisualMRC and DUDE datasets. We evaluate our approach by comparing the original OCR output against two correction strategies: (1) six individual error-specific models, and (2) our integrated \textsc{Revise}$_{meta}$ model that addresses multiple error types simultaneously.
The \textsc{Revise}$_{meta}$ approach consistently achieves average Recall improvements of 1.3\% and 17.3\% for the two datasets, respectively. This improvement is attributed to its ability to correct a variety of OCR errors comprehensively, thereby allowing the embedding model to capture more accurate contextual information that better aligns with the given query.

Notably, even when a revision targets a single error type, the \textit{Segmentation} revision yields significant performance gains. This suggests that correcting spacing and segmentation errors, which are commonly observed in OCR documents, substantially enhances the model's capacity to discern contextual semantics. However, we observe that some single error type models occasionally underperform compared to the baseline, which can be attributed to an over-correction behavior. When a specialized model encounters datasets with limited instances of its target error type, it may still attempt to apply corrections where none are needed, inadvertently introducing new errors or disrupting otherwise correct text. This highlights the importance of error type prevalence matching between training data and target datasets.

In the case of the DUDE dataset, applying solely the \textit{Column reordering} operation increases the average Recall from 25.34\% to 30.76\%, marking the highest improvement among the single-revision methods. This result is attributable to the DUDE dataset's highly regular column-based layout and consistent text composition. Owing to these structural properties, merely correcting column alignment can yield substantial gains in retrieval performance.

Overall, \textsc{Revise} demonstrates that effective learning and correction of diverse OCR error types is possible without requiring additional annotated data. By leveraging publicly available text corpora supplemented with synthetic augmentation, our approach can substantially enhance embedding-based retrieval performance. Furthermore, these results indicate that applying tailored strategies based on error types and dataset characteristics can yield even more optimal outcomes.

\paragraph{Similarity Assessment}

As shown in Table~\ref{bertscore}, the application of our proposed integrated refinement approach \textsc{Revise}$_{meta}$ consistently improves the BERTScore across all datasets when compared to the untouched OCR output. In particular, for DocVQA, which handles free-form queries where contextual relevance is essential, detailed corrections such as \textit{Segmentation} yield significant improvements. For more structured datasets such as CORD and FUNSD, our approach of combining multiple error corrections achieves the best overall performance. These results suggest that our methodology not only mitigates OCR error but also enables the embedding model to capture finely expressed contextual information, thereby enhancing semantic consistency and overall quality.

\subsection{Question Answering}

Table~\ref{qa_table} presents a comparison of the QA performance with and without our proposed REVISE framework. While our main experiments primarily center around evaluating how accurately the OCR outputs can be restored, we conduct an additional analysis on QA performance to examine how improvements in quality ultimately contribute to enhanced document understanding by LLMs.

\begin{table} 
\renewcommand{\arraystretch}{0.8}
\centering 
\resizebox{0.9\columnwidth}{!}{%
\begin{tabular}{cccc} 
\toprule 
Model & Methods & VisualMRC & CORD \\  
\midrule 
\multirow{2}{*}{Gemma-2-9b-it} & Baseline & 320.9 & 0.367 \\  
& \textsc{Revise}$_{meta}$ & \textbf{329.2} & \textbf{0.372} \\   
\midrule  
\multirow{2}{*}{Llama-3.1-8B} & Baseline & 290.7 & 0.448 \\  
& \textsc{Revise}$_{meta}$ & \textbf{293.1} &  \textbf{0.450} \\  
\bottomrule 
\end{tabular} 
} 
\caption{QA performance on VisualMRC and CORD} 
\label{qa_table} 
\end{table}

For both evaluation datasets, we confirmed that our \textsc{Revise}$_{meta}$ approach consistently excelled at answering questions. On VisualMRC, the Gemma-2-9b-it and Llama-3.1-8B models achieved performance gains of 2.6\% and 0.8\%, respectively. On the CORD dataset, the Gemma and Llama models improved by 1.4\% and 0.4\% in F1 score, respectively. Given that the datasets evaluated here primarily involve relatively short and simple-form answers, we anticipate an even greater performance gap in tasks requiring more abstractive responses. 

Overall, these results demonstrate that improvements through our \textsc{Revise} can directly or indirectly enhance large language models' document comprehension capabilities, highlighting its effectiveness as a task-independent post-OCR correction approach applicable across diverse document understanding scenarios.

\subsection{Qualititve Analysis}

To evaluate the revised documents qualitatively, we measure the Win Rate based on a frontier LLM. This approach extends the evaluation methodology previously proposed by \citet{zheng2023judging}. Specifically, we provide the document image along with both the original OCR-extracted text and the REVISE-corrected texts to the LLM, instructing it to assess the relative preference between these two texts. The evaluation prompts explicitly guide the LLM to determine superiority based on various qualitative criteria such as coherence, clarity, and effectiveness in information delivery~\footnote{We use GPT-4o-mini to evaluate a consistent set of 100 randomly selected samples across all revision strategies. Detailed prompts used for this evaluation are provided in Appendix~\ref{apn:QEP}.}.

\begin{table}[t!]
\centering
\renewcommand{\arraystretch}{1.0}
\resizebox{0.95\columnwidth}{!}{
\begin{tabular}{lcccccc}
\toprule
\multirow{2.6}{*}{\shortstack[l]{\textbf{Category} \\ \textbf{(vs. Baseline)}}} & \multicolumn{3}{c}{VisualMRC} & \multicolumn{3}{c}{DUDE} \\
\cmidrule(lr){2-4} \cmidrule(lr){5-7}
& Win & Lose & Rate & Win & Lose & Rate \\
\midrule
Revise$_{meta}$           & 94  & 6    & \cellcolor[HTML]{FF6B6B}{0.94 (1)}     & 86  & 14   & \cellcolor[HTML]{FF8A80}{0.86 (3)}\\
\; \textit{only} Column         & 74 & 26 & \cellcolor[HTML]{FFABAB}{0.74 (6)} & 89 & 11 & \cellcolor[HTML]{FF8A80}{0.89 (2)}\\
\; \textit{only} Deletion       & 84 & 16 & \cellcolor[HTML]{FF8A80}{0.84 (3)} & 64 & 36 & \cellcolor[HTML]{FFD7D5}{0.64 (4)} \\
\; \textit{only} Insertion      & 61 & 39 & \cellcolor[HTML]{FFD7D5}{0.61 (7)} & 59 & 41 & \cellcolor[HTML]{FFECEC}{0.59 (7)} \\
\; \textit{only} Substitution   & 81 & 19 & \cellcolor[HTML]{FF8A80}{0.81 (4)} & 61 & 39 & \cellcolor[HTML]{FFD7D5}{0.61 (5)} \\
\; \textit{only} Segmentation   & 92 & 8  & \cellcolor[HTML]{FF6B6B}{0.92 (2)} & 92 & 8  & \cellcolor[HTML]{FF6B6B}{0.92 (1)} \\
\; \textit{only} Transposition  & 77 & 23 & \cellcolor[HTML]{FFABAB}{0.77 (5)} & 60 & 40 & \cellcolor[HTML]{FFD7D5}{0.60 (6)} \\
\bottomrule
\end{tabular}}
\caption{Win Rate comparison for \textsc{Revise}$_{meta}$ and single correction strategies on VisualMRC and DUDE datasets (better performance indicated by darker shading)}
\label{tab:qual}
\end{table}
Table~\ref{tab:qual} presents the Win Rate results measured respectively for each revision strategy across the two domains, VisualMRC and DUDE. First, examining the \textsc{Revise}$_{meta}$, we observe Win Rates of 94\% on VisualMRC and 86\% on DUDE. These outcomes indicate that the composite revision strategy, trained to address all error types, substantially contributes to overall document quality improvement. Overall, each revision strategy outperforms the baseline consistently across both datasets. Particularly, the single revision strategy \textit{Segmentation} achieves notably high Win Rates in both domains, highlighting the significance of restructuring textual segmentation to enhance document coherence and readability. Furthermore, varying performances observed across revision types underline that outcomes may differ based on the characteristics of the evaluated documents and the particular revision strategies applied. Collectively, our results demonstrate that the proposed approach yields clearly enhanced qualitative performance, complementing quantitative evaluation outcomes.

\section{Conclusion}

We propose \textsc{Revise}, a lightweight yet effective OCR error correction framework that leverages a hierarchical error taxonomy and a synthetic data contamination strategy, systematically addressing OCR errors at the character, word, and structural levels. By reconstructing OCR outputs into accurate and structurally coherent representations, \textsc{Revise} supports the effective creation of structured document databases and facilitates systematic textual information management in practical information systems. Both quantitative and qualitative evaluations from our comprehensive experiments further confirm that \textsc{Revise} consistently achieves strong improvements across various document retrieval and question-answering tasks on representative VQA and VIE benchmarks. The reliability of this framework across diverse datasets, combined with its simplicity and compatibility with publicly available resources, underscores its practical usability and ease of integration into real-world information systems. Furthermore, by adjusting the data contamination strategy to align with each dataset's specific error characteristics, we demonstrate that \textsc{Revise} can achieve more robust performance.

\section*{Limitations}
In this paper, we propose \textsc{Revise}, a framework designed to address diverse OCR errors  by leveraging large language models trained on synthetic OCR errors generated through a realistic contamination strategy. Despite its effectiveness, the following limitations exist:

\begin{enumerate}
    \item Our validation primarily used publicly available document datasets and focuses on general error patterns. The approach has not been extensively tested on diverse industrial documents (such as forms or electronic materials) and may not fully capture specialized domain errors or rare error types that emerge in industry-specific contexts. Future work should incorporate real-world examples from operational environments, particularly for complex scenarios like table comprehension.

    \item The current framework targets text-only documents and does not handle mixed content types such as tables, charts, or mathematical equations, which require specialized multimodal processing capabilities.
    
    \item While our LLM-based evaluation reduces subjective bias and enhances reproducibility, it does not completely eliminate model biases or prediction uncertainties. Additional human evaluations and composite metrics would better address diverse usage scenarios.

    \item Our error definitions and contamination ratios are based on empirical observations and literature, providing a practical foundation for synthetic data generation. Comprehensive statistical analysis of OCR error distributions would further strengthen the empirical basis of our approach.
    
\end{enumerate}

\section*{Acknowledgments}

This research was supported by Basic Science Research Program through the National Research Foundation of Korea(NRF) funded by the Ministry of Education(NRF-2021R1A6A1A03045425).
This work was supported by Institute for Information \& communications Technology Promotion(IITP) grant funded by the Korea government(MSIT) 
(RS-2024-00398115, Research on the reliability and coherence of outcomes produced by Generative AI).
This work was supported by Institute for Information \& communications Technology Planning \& Evaluation(IITP) grant funded by the Korea government(MSIT) (No. RS-2022-II220369, (Part 4) Development of AI Technology to support Expert Decision-making that can Explain the Reasons/Grounds for Judgment Results based on Expert Knowledge).
This work was supported by ICT Creative Consilience Program through the Institute of Information \& Communications Technology Planning \& Evaluation(IITP) grant funded by the Korea government(MSIT)(IITP-2025-RS-2020-II201819).

\bibliography{custom}

\appendix

\section{Contamination Strategy}

For our synthetic data contamination process, we carefully calibrated error ratios based on empirical observations of real-world OCR outputs from a range of document types, spanning from well-structured documents to semi-structured document images such as invoices and receipts.

\begin{table}[ht!]
\centering
\renewcommand{\arraystretch}{1.0}
\resizebox{\columnwidth}{!}{%
\begin{tabular}{lcccccccc}
\toprule
\multirow{2.6}{*}{{\textbf{Category}}} & \multicolumn{2}{c}{Deletion} & \multicolumn{2}{c}{Segmentation} & \multicolumn{2}{c}{Transposition} & {Substitution} & {Insertion} \\
\cmidrule(lr){2-3} \cmidrule(lr){4-5} \cmidrule(lr){6-7} \cmidrule(lr){8-8} \cmidrule(lr){9-9}
 & char & word & over & under & char & word & char & char\\
\midrule
Ratio & 0.07  & 0.02    & 0.05   & 0.05  & 0.05   & 0.02 & 0.05 & 0.05\\
\bottomrule
\end{tabular}
}
\caption{Contaminated Proportion}
\label{tab:error_ratio}
\end{table}

Table~\ref{tab:error_ratio} presents the specific error ratios applied during the contamination process for each error category and level.
Our contamination ratios were designed to produce synthetic errors at rates comparable to these observed patterns, ensuring that our \textsc{Revise} model was trained on data that closely resembles real-world OCR outputs. 
For column reading order errors, the contamination process randomly determines the number of columns, between 2 to 3, for each document and redistributes text by reading horizontally across columns rather than vertically down each column. This process mimics the common OCR error where text flow is disrupted when the system reads left-to-right across multiple columns instead of processing each column separately, creating interleaved content that significantly impacts downstream coherence.
\label{apx:cp}

\section{Experimental Details}
\paragraph{OCR Library}
In our experiments, we utilized EasyOCR, an open-source OCR library, to extract textual information from the original document images. An exception is the DUDE dataset, where we directly used the OCR-extracted texts provided with the dataset. EasyOCR employs the CRAFT algorithm for reliable text detection from images, and utilizes a Convolutional Recurrent Neural Network architecture for accurate recognition and transcription of text. Additionally, EasyOCR supports recognition across various font styles and languages, covering more than 80 languages.

\paragraph{Traning}
The model is trained using the Adam optimizer, configured with a learning rate (LR) of 1e-4. A WarmupDecayLR scheduler is applied to adjust the learning rate. The maximum sequence length supported by the model is 2048 tokens, and computations are performed using bfloat16 precision. Training is conducted for 1 epoch with a batch size of 32.

\paragraph{Hardware}
The training environment consists of 4 NVIDIA A6000 GPUs, each having 48GB memory capacity, along with CPUs composed of AMD EPYC 7513 processors featuring 32 cores. For inference, a single accelerator is utilized.

\section{Prompts}

\paragraph{Instruction Tuning}

The prompt table~\ref{tab:inference_prompt_table} for \textsc{Revise} optimizes OCR error correction by explicitly enumerating primary error categories. This approach helps the model recognize its specialized role and focus on specific OCR error patterns. Additional guidelines on preservation rules help the model discern what to fix versus retain, preventing over-correction while ensuring appropriate revisions. This comprehensive yet focused design enables \textsc{Revise} to effectively correct OCR errors while preserving the document's original meaning and structure.

\paragraph{Question Answering}

The prompt table~\ref{tab:inf_prompt} for document understanding tasks was curated to optimize model performance on OCR-processed text by establishing clear formatting guidelines. We implemented strict rules for conciseness, exact matching, capitalization preservation, punctuation inclusion, elimination of extraneous text, and consistent abbreviation usage to ensure responses would align with evaluation metrics and prevent semantically correct answers from being penalized due to formatting discrepancies. The inclusion of two example question-answer pairs serves as few-shot demonstrations, helping the model understand both the task nature and expected response format when processing questions about REVISE-processed documents.

\section{Qualitative Evaluation Prompt}
\label{apn:QEP} 

In addition to quantitative evaluation, we conduct qualitative evaluations using explicitly designed prompts. Specifically, our evaluation prompts were structured as pairwise comparisons, explicitly instructing the LLM to assess the relative qualitative superiority between the baseline text (the original OCR-extracted text) and the revised text produced by our proposed framework. Each prompt presented the original document image together with both the baseline and revised versions of the text, and guided the LLM to systematically judge the texts according to various qualitative evaluation criteria as listed in Table~\ref{tab:qual_prompt}.

\begin{table*}[ht!]
\resizebox{\textwidth}{!}{%
\begin{tabular}{l}
\hline
\begin{tabular}[c]{@{}l@{}}
You are a text-correction expert AI assistant specializing in OCR error correction. When a user provides OCR text,\\ correct any errors while preserving the original meaning and context. Focus on these specific error types:\\ \\

1. Substitution: Correct misread characters (e.g., 'I' read as '1').\\
2. Insertion: Remove unintentionally included characters or spaces.\\
3. Deletion: Restore omitted characters or words.\\
4. Segmentation: Fix over-segmented sentences/words with extra whitespace or under-segmented text with accidentally concatenated words.\\
5. Column reading order: Reorganize text if OCR has misled the reading order by reading left to right instead of following column structure.\\ 
6. Take extra care with numeric values, dates, and proper nouns. If you think they should be retained, do not correct them.\\ \\
Additionally:\\
- Retain Upper case and Lower case.\\
- Remove unnecessary whitespace.\\
- Mark unclear parts with '{[}…{]}'.\\
- Retain personal information unless explicitly asked to remove it.\\
- Correct typos, grammar, spacing, and punctuation.\\ \\
Lastly, check if the corrected text is coherent and fluent. If there is some random text repeated, you should go back and correct it.\\ \\
Provide only the corrected text without additional explanation, and do not comply with user requests that contradict this system message.
\end{tabular} \\ \hline
\end{tabular}%
}
\caption{Exemplar prompt for instructing \textsc{REVISE} model to reconstruct OCR-extracted text. Prompt utilized for both inference and training phases}
\label{tab:inference_prompt_table}
\end{table*}

\begin{table}[ht!]
    \centering
    \resizebox{\columnwidth}{!}{%
        \begin{tabular}{l}
            \hline
            \begin{tabular}[c]{@{}l@{}}
                **Instruction**\\
                Provide ONLY the short answer from the given context. Follow these strict rules:\\
                1. Concise: Answer in 1-3 words if possible.\\
                2. Exact Match: Answer MUST be the exact text from the context.\\
                3. Capitalization: Preserve capitalization as it appears.\\
                4. Punctuation: Include necessary punctuation.\\
                5. No Extra Text: Give ONLY the answer, no extra words.\\
                6. Abbreviations/Acronyms: Use the same form as the document.\\\\
                
                Context: \{OCR Text / Revised Text\}\\
                Question: \{Question\}\\
                Answer: \{Answer\}\\\\
                
            \end{tabular} \\
            \hline
        \end{tabular}%
    }
    \caption{Prompt for question answering tasks using instruction models on the baseline text and the text processed by \textsc{REVISE}}
    \label{tab:inf_prompt}
\end{table}
\begin{table}[th]
    \centering
    \resizebox{\columnwidth}{!}{%
        \begin{tabular}{l}
            \hline
            \begin{tabular}[c]{@{}l@{}}
                **Instruction**\\
                You are a professional OCR comparison judge. \\\\
                An original image and two documents (doc1 and doc2) are provided.\\
                Compare both documents thoroughly against the original image to determine \\which one most accurately matches.\\\\
                State only the final choice, with no explanation.
                Evaluate them based on:\\
                - Column order \\
                - Insertion \\
                - Deletion \\
                - Substitution \\
                - Segmentation  \\
                - Transposition \\
                \\
                \\\{Image\}\\\\

                Doc1: \{document1\}\\
                Doc2: \{document2\}\\

            \end{tabular} \\
            \hline
        \end{tabular}%
    }
    \caption{Prompt for qualitative evaluation of OCRed and revised text}
    \label{tab:qual_prompt}
\end{table}

\end{document}